\documentclass{article}

\usepackage{arxiv}
\usepackage{authblk}
\usepackage[utf8]{inputenc}
\usepackage[T1]{fontenc}
\usepackage{hyperref}
\usepackage{url}
\usepackage[round]{natbib}  
\usepackage{booktabs}
\usepackage{amsfonts}
\usepackage{nicefrac}
\usepackage{microtype}
\usepackage{graphicx}
\usepackage{xcolor}
\graphicspath{{./image/}}

\usepackage{bm}
\usepackage{amsmath,amssymb}
\usepackage{amsthm}
\usepackage[shortlabels]{enumitem}
\usepackage{siunitx}
\usepackage{listings}
\usepackage{float}
\usepackage{placeins}

\theoremstyle{definition}

\title{Emulating Retrieval Augmented Generation via Prompt Engineering
for Enhanced Long Context Comprehension in LLMs}

\author{Joon Park}
\author{Kyohei Atarashi}
\author{Koh Takeuchi}
\author{Hisashi Kashima}
\affil{Kyoto University}

\begin{document}
\maketitle

\begin{abstract}
This paper addresses the challenge of comprehending very long contexts in Large Language Models (LLMs) by proposing a method that emulates Retrieval Augmented Generation (RAG) through specialized prompt engineering and chain-of-thought (CoT) reasoning. While recent LLMs support over 100,000 tokens in a single prompt, simply enlarging context windows has not guaranteed robust multi-hop reasoning when key details are scattered across massive input. Our approach treats the model as both the retriever and the reasoner: it first tags relevant segments within a long passage, then employs a stepwise CoT workflow to integrate these pieces of evidence. This single-pass method thereby reduces reliance on an external retriever, yet maintains focus on crucial segments.
We evaluate our approach on selected tasks from BABILong, which interleaves standard bAbI QA problems with large amounts of distractor text. Compared to baseline (no retrieval) and naive RAG pipelines, our approach more accurately handles multi-fact questions such as object location tracking, counting, and indefinite knowledge. Furthermore, we analyze how prompt structure, including the order of question, relevant-text tags, and overall instructions, significantly affects performance. These findings underscore that optimized prompt engineering, combined with guided reasoning, can enhance LLMs' long-context comprehension and serve as a lightweight alternative to traditional retrieval pipelines.
\end{abstract}

\section{Introduction}
\label{sec:intro}
Large Language Models (LLMs) have demonstrated remarkable progress in understanding and generating human language, scaling from initial architectures capable of handling a few hundred tokens to recent systems supporting context windows exceeding 100{,}000 tokens. Such expansions in context length enable LLMs to process extensive documents---including entire novels, legal contracts, or scientific reports---in a single prompt. Despite this progress, LLMs still face critical challenges when dealing with \emph{very long} inputs containing dispersed, multi-faceted information \citep{kuratov2024babilong, li2024needlebench, wang2024multimodal}. In particular, simply increasing context size does not guarantee that a model will accurately retrieve and combine distant pieces of information. Models often fail on tasks requiring multi-hop reasoning, especially when relevant details are scattered across large portions of text \citep{lee2024can, adams2024longhealth, karpinska2024one, levy-etal-2024-task}.

One promising line of research has focused on \emph{Retrieval-Augmented Generation (RAG)}, in which an LLM is augmented with an external retrieval mechanism to fetch relevant passages from a massive corpus or knowledge base. This approach efficiently narrows down the input, enabling the model to focus on a set of shorter, contextually relevant documents \citep{reddy-etal-2024-docfinqa}. While RAG excels at pinpointing factual snippets in long contexts, it often struggles with \emph{multi-hop} or \emph{compositional} reasoning: if multiple evidence fragments must be pieced together, a naive retrieve-then-summarize workflow can fail to integrate that information cohesively \citep{bai-etal-2024-longbench}.

In parallel, \emph{Chain-of-Thought (CoT)} prompting has emerged as a powerful technique for guiding LLMs through explicit intermediate reasoning steps \citep{wei2022chain}. By illustrating a sequence of thought processes in the prompt, CoT enables the model to better break down a question, isolate the key aspects of a problem, and infer the correct answer. However, when crucial facts are omitted or buried in large swaths of irrelevant content, CoT alone may lead to hallucinated or incomplete reasoning \citep{agarwal2024manyshot}. Therefore, a synergistic approach is needed that combines the coverage advantages of RAG with the fine-grained, step-by-step problem solving of CoT.

In this research, we propose a novel method that \textbf{emulates RAG through prompt engineering and chain-of-thought reasoning}, aiming to enhance LLM capabilities in long-context comprehension while mitigating the drawbacks of standard retrieval. Our approach treats the model as both the retriever and the reasoner: first identifying relevant segments within a large context, then injecting explicit reasoning traces to simulate multi-hop retrieval. By doing so, we attempt to unify the best of both worlds: (1) \emph{focus} on relevant information scattered throughout the text, and (2) \emph{stepwise reasoning} to combine evidence into coherent answers.

We test our method on the BABILong dataset \citep{kuratov2024babilong} requiring advanced reasoning over thousands of tokens, including large-narrative benchmarks such as NovelQA \citep{wang2025novelqa}. Experimental results indicate that our approach can outperform standard RAG pipelines on certain multi-fact tasks while also alleviating context limitations for smaller-parameter models. In particular, we demonstrate improvements on counting, indefinite knowledge questions, and multi-step object tracking within extensive textual inputs. Moreover, we compare our approach against vanilla long-context models and observe that dedicated retrieval--reasoning loops yield more robust integration of distant details.

The remainder of this paper is organized as follows. In Section~\ref{sec:background}, we present an overview of related work, including retrieval-augmented approaches, chain-of-thought prompting, and benchmarks for long-context LLMs. Section~\ref{sec:proposed} describes our proposed method in detail, focusing on how we guide the model to identify and utilize relevant context segments. Section~\ref{sec:experiments} outlines our experimental setup, datasets, and implementation details. We then report and analyze the results---including detailed per-task findings and the impact of different prompt orders---in Section~\ref{sec:results_and_analysis}. Finally, we conclude and discuss future directions in Section~\ref{sec:conclusion}.

\section{Background and Related Work}
\label{sec:background}

\subsection{Evolution of Long Context in LLMs}
Early language models (e.g., BERT, GPT-2) were constrained by input lengths of a few hundred to a couple thousand tokens. More recent models have dramatically increased their capacity: GPT-3 scaled to 2{,}048 tokens, and subsequent architectures, such as GPT-3.5 Turbo, Claude 2, and Gemini 1.5, now claim support for input contexts beyond 100{,}000 tokens \citep{georgiev2024gemini}. Nevertheless, empirical studies reveal that these long-context capabilities do not necessarily translate to effective \emph{utilization} of the entire input. Despite large or even multimillion-token windows, models often attend to only a fraction of the provided text, leading to performance degradation on reasoning tasks where critical information is scattered far apart \citep{kuratov2024babilong, zhang2024infinitybench, levy-etal-2024-task}.

Multiple benchmarks have emerged to evaluate how well LLMs leverage extended context. For instance, \cite{kuratov2024babilong} introduced \emph{BABILong}, which provides tasks specifically designed to test needle-in-a-haystack retrieval and multi-step reasoning within extensive inputs. Similarly, \cite{li2024needlebench} presented \emph{NeedleBench}, extending context lengths up to one million tokens to assess how models handle single-needle or multi-needle retrieval challenges. Other datasets like \emph{NovelQA} \citep{wang2025novelqa}, \emph{One Thousand and One Pairs} \citep{karpinska2024one}, and \emph{CLongEval} \citep{qiu-etal-2024-clongeval} test performance on massive text corpora, including entire novels and Chinese-language benchmarks. Summarization techniques for large texts can be considered complementary or integrated with retrieval \citep{koh2022empirical}.

The overarching conclusion from these works is that \emph{merely expanding context size} does not guarantee robust comprehension and multi-fact integration. Key issues include \emph{context confusion} and \emph{context dilution} in the attention mechanism, making it difficult for the model to prioritize and recall crucial facts \citep{lee2024can}. Consequently, the quest for effective \emph{long-context understanding} has shifted toward techniques such as retrieval-based solutions, chunking, hierarchical modeling, or refined prompt engineering to handle large text inputs \citep{shaham2022scrolls}.

\subsection{Retrieval-Augmented Generation (RAG)}
\label{subsec:rag}
One widely adopted strategy is \emph{Retrieval-Augmented Generation (RAG)}, where a language model is paired with an external retriever that fetches relevant snippets from an indexed corpus \citep{lee2024can, reddy-etal-2024-docfinqa}. Rather than providing the entire document, the retriever selects a small subset of text presumably related to the user's query. While RAG effectively narrows the attention to relevant content, it typically treats retrieval as a one-shot step \citep{bai-etal-2024-longbench}. Multi-hop reasoning tasks requiring multiple disjoint facts can falter if the initial top-$k$ retrieval does not capture all necessary segments \citep{li2024needlebench}.

Various iterative or tool-augmented retrieval methods have emerged to address these limitations, interleaving generation with refined retrieval calls \citep{lee2024can}. However, these multi-round pipelines add complexity and computational overhead. Our approach aims to preserve multi-hop \emph{retrieval} within a single pass of the model (see Section~\ref{sec:proposed}), thus offering a lightweight alternative to full-scale external RAG.

\subsection{Chain-of-Thought Reasoning}
\label{subsec:cot}
\emph{Chain-of-thought (CoT) prompting} has shown that LLMs can significantly improve on complex tasks by producing intermediate reasoning steps \citep{wei2022chain}. However, when crucial facts are missing or buried in large contexts, CoT by itself may lead to incomplete or hallucinated chains of reasoning \citep{agarwal2024manyshot}. Consequently, an important line of work explores combining CoT with external retrieval calls \citep{shaham2022scrolls, lee2024can}. Yet such an approach can require multiple model passes or a separate retriever.

In this paper, we incorporate CoT reasoning \emph{within} a single forward pass that also emulates retrieval via prompt engineering. Rather than calling an external index, we ask the model to identify relevant text segments in the prompt itself, then chain those segments into a multi-hop solution.

\subsection{Emulating RAG via Prompt Engineering}
\label{subsec:emulating_rag}
The main idea behind \emph{emulating RAG} is to unify the benefits of retrieval-based focusing and CoT-based multi-step reasoning within a single prompt. Instead of orchestrating separate retrieval calls, we instruct the model to \emph{locate and tag} relevant portions of the input text, then walk through these tagged segments in a CoT style \citep{li2024needlebench}.

This approach hinges on carefully structured prompts (Section~\ref{sec:proposed}), where the language model is guided to:
\begin{enumerate}
    \item Identify relevant snippets.
    \item Tag them explicitly (e.g., with \texttt{<relevant\_section>}).
    \item Summarize or analyze those snippets in a chain-of-thought manner.
    \item Produce the final answer, referencing the tagged evidence.
\end{enumerate}
By encapsulating retrieval-like selection and multi-hop reasoning in one pass, we reduce reliance on external retrievers while retaining robust coverage of potentially large input contexts.

\section{Proposed Method}
\label{sec:proposed}

In this section, we detail our method for \emph{prompt-based retrieval emulation} and \emph{chain-of-thought (CoT)} reasoning, offering a more in-depth discussion of the workflow and design choices than prior brief overviews. Figure~\ref{fig:method-overview} gives a schematic, which we expand upon below.

\subsection{Overall Architecture}
Our approach aims to reproduce the key steps of Retrieval-Augmented Generation (RAG) which include retrieval and subsequent generation within a single large language model (LLM) call. Traditional RAG pipelines typically rely on:
\begin{enumerate}
    \item A separate retrieval system (e.g., a vector search engine),
    \item A short list of fetched passages,
    \item A final generation step conditioned on those passages.
\end{enumerate}
By contrast, our method leverages \emph{prompt engineering} to simulate retrieval: the model is instructed to (a) mark relevant sections in the text, (b) reference them explicitly, and (c) combine them via chain-of-thought reasoning. This unification seeks to minimize overhead, as we avoid repeated calls to external search engines.

\begin{figure}[t]
\centering
\includegraphics[width=0.95\textwidth]{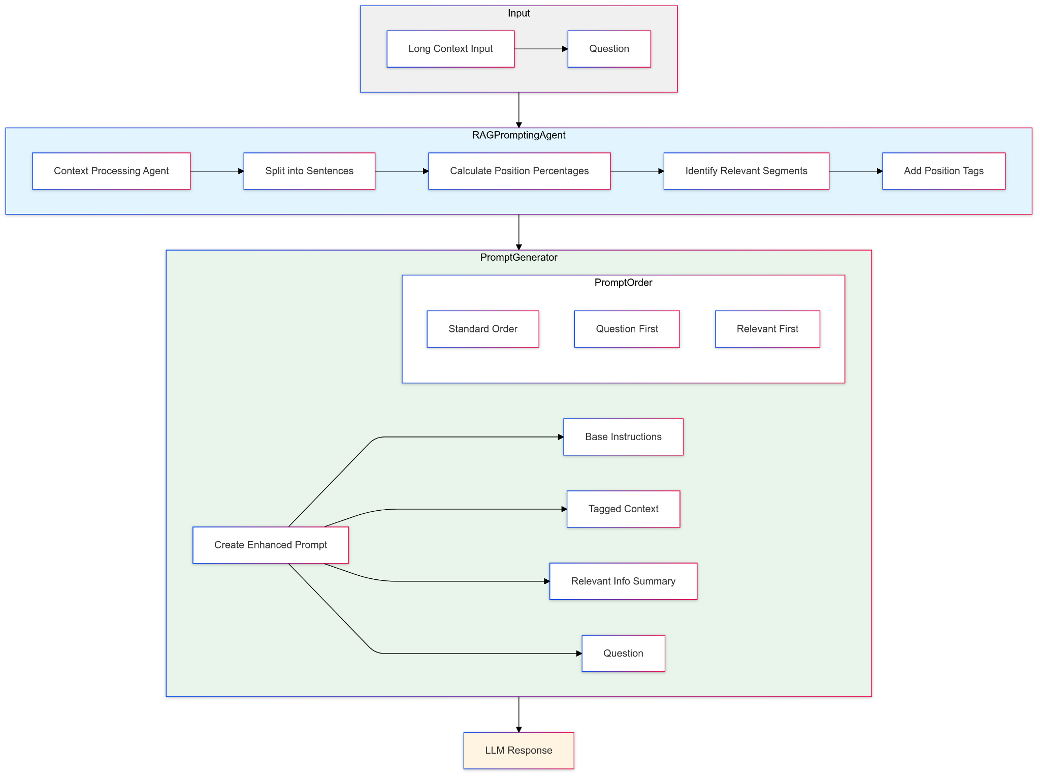}
\caption{Pipeline of Our Proposed Method}
\label{fig:method-overview}
\end{figure}

As illustrated in Figure~\ref{fig:method-overview}, the method proceeds in a single forward pass:

\begin{enumerate}
    \item \textbf{Tagging / Relevance Detection (Internal Retriever).} 
    \item \textbf{Localized Summaries / Extraction.} 
    \item \textbf{Chain-of-Thought Reasoning / Synthesis.} 
    \item \textbf{Final Answer Generation.} 
\end{enumerate}

Below, we delve into each step in more detail.

\subsection{Step 1: Tagging / Relevance Detection}
\label{subsec:step1_tagging}
In typical retrieval pipelines, a query is used to find relevant paragraphs or documents. In our single-pass approach, we embed instructions that direct the model to look for content relevant to a certain query or question \emph{within} the prompt. For instance, we might format the prompt like this (illustrative example):

\begin{quote}
\small
\texttt{[INSTRUCTION PART]:}\\
\textit{You will be given a very long passage. First, identify any sentences or paragraphs that are relevant to the question. Surround each relevant snippet in a \texttt{<relevant\_section>} tag. Also note the approximate position.}\\

\texttt{[CONTENT]:}\\
(Here is a very long text, thousands of tokens)...

\texttt{[QUESTION]:}\\
"What is the final location of the apple?"
\end{quote}

The LLM, if well-instructed and sufficiently capable, will output text that includes tags like:
\begin{verbatim}
<relevant_section position="12.3%">
   Daniel grabbed the apple in the garden
</relevant_section>
...
<relevant_section position="58.9%">
   Daniel moved to the kitchen
</relevant_section>
\end{verbatim}
This effectively simulates retrieval by isolating relevant segments. We emphasize that the model's capacity to follow these instructions depends on its instruction-following and long-context comprehension skills.

\paragraph{Granularity and Format.}
We can vary granularity (e.g., sentence-level or paragraph-level tagging) and specify a JSON-based format if more structure is desired. For example, we can ask for a JSON array named \texttt{relevant\_sections}, each containing \texttt{content}, \texttt{position}, and \texttt{reason\_why\_tagged}. This portion of the prompt encourages the model to self-rationalize its tagging, potentially revealing how it interprets the question.

\paragraph{Why It Works (If It Does).}
Models with strong instruction-following capabilities can parse the question, infer which parts of the text matter, and faithfully tag them. This stands in place of a separate search index. However, success here is not guaranteed if the LLM is too small or the text is overly complex. In practice, our experiments on BABILong show that with careful prompt design, modern LLMs can reliably identify relevant segments even in tens of thousands of tokens.

\subsection{Step 2: Localized Summaries / Extraction}
\label{subsec:step2_localization}
After tagging, we can prompt the model (in the \emph{same output}) to generate a localized summary or extraction for each \texttt{<relevant\_section>}. For instance, the instructions might read:
\begin{quote}
\small
\texttt{Now, for each relevant section you tagged, please summarize any entities, actions, or numeric details, storing them in a short bullet list.}
\end{quote}
This step can help prevent losing important context. In a multi-hop query, we might have multiple relevant sections scattered across different portions of the text. Summaries provide a condensed version of each segment, ready for integration in the next step. 

\subsection{Step 3: Chain-of-Thought Reasoning / Multi-Hop Integration}
\label{subsec:step3_cot}
Next, the model is prompted (still within the same single output sequence) to \emph{combine} these localized summaries and produce a chain-of-thought explanation. Because we have summaries from each relevant section, the model can do multi-hop reasoning. For a location-tracking question, the reasoning might look like:

\begin{quote}
\small
\textbf{Step 1:} Daniel initially grabbed the apple in the garden (position 12.3\%).\\
\textbf{Step 2:} Daniel moved to the kitchen (position 58.9\%).\\
\textbf{Step 3:} Therefore, the apple is now in the kitchen.
\end{quote}

\subsection{Step 4: Final Answer Generation}
\label{subsec:step4_answer}
Finally, after enumerating the chain-of-thought steps, the model outputs a concise result:
\begin{quote}
\small
\textbf{Answer:} The apple is in the kitchen.
\end{quote}
In tasks requiring “yes/no/maybe” or a numeric response, the final statement can adhere to the required format. By appending a clear directive (e.g., “Now provide the final answer in one sentence”), we help the model produce a succinct, well-structured conclusion.

\subsection{Implementation Details and Prompt Organization}
\label{subsec:prompt_details}
\begin{figure}[t]
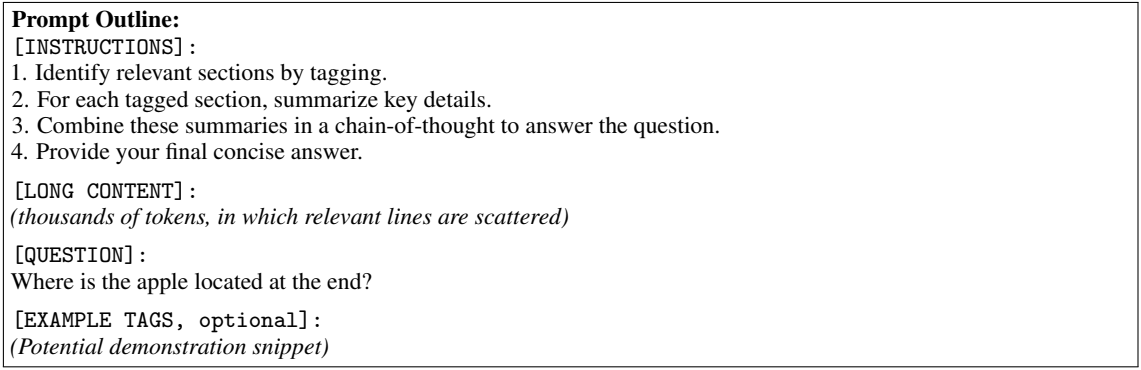

\centering
\fbox{\begin{minipage}{0.9\textwidth}
\footnotesize
\textbf{Prompt Outline:}\\
\texttt{[INSTRUCTIONS]:}\\
1. Identify relevant sections by tagging.\\
2. For each tagged section, summarize key details.\\
3. Combine these summaries in a chain-of-thought to answer the question.\\
4. Provide your final concise answer.

\vspace{0.5em}
\texttt{[LONG CONTENT]:}\\
\textit{(thousands of tokens, in which relevant lines are scattered)}

\vspace{0.5em}
\texttt{[QUESTION]:}\\
Where is the apple located at the end?

\vspace{0.5em}
\texttt{[EXAMPLE TAGS, optional]:}\\
\textit{(Potential demonstration snippet)}

\end{minipage}}
\caption{Illustrative Prompt Construction for Single-Pass Retrieval Emulation + CoT}
\label{fig:expanded_prompt}
\end{figure}

We have described four conceptual steps, but crucially, we implement them in a \emph{single} forward pass. That is, we do not make multiple calls to the LLM. Instead, the prompt is carefully organized to \emph{ask} the model to produce tagged segments, local summaries, chain-of-thought, and final answer \emph{in sequence} within one completion. Figure~\ref{fig:expanded_prompt} outlines a simplified version of how the final prompt might look.

\paragraph{Prompt Order Variations.}
As later described in Section~\ref{sec:results_and_analysis}, we experimented with different orders of presenting the question and relevant information to the model. The exact structure of the final prompt can significantly influence performance on multi-hop tasks.

\paragraph{Advantages.}
\begin{itemize}
    \item \textbf{Reduced Overhead:} We avoid repeated calls to a separate retriever or multiple LLM passes.
    \item \textbf{Integrated Explanation:} The chain-of-thought mechanism naturally merges retrieval-like tagging with multi-hop reasoning.
    \item \textbf{Flexible Format:} Users can refine the instructions (e.g., requiring bullet-point summaries, JSON objects, or references to snippet positions).
\end{itemize}

\paragraph{Limitations.}
\begin{itemize}
    \item \textbf{Reliance on Model Obedience:} If the LLM does not reliably follow instructions, the method can fail (e.g., missing or incorrect tagging).
    \item \textbf{Large Prompt Size:} We must fit the entire text plus instructions and chain-of-thought in one prompt, which can be challenging if the text is extremely large.
    \item \textbf{Single-Pass Boundaries:} True iterative retrieval---where new evidence might alter the search query---is not directly replicated. If a follow-up question arises, a new forward pass is needed.
\end{itemize}

\section{Experiments}
\label{sec:experiments}
In this section, we describe our experimental setup, implementation details, and evaluation protocols. We first introduce the BABILong dataset in Section~\ref{subsec:data}, then present our models, baselines, and the ways we measure performance.

\subsection{Dataset: BABILong}
\label{subsec:data}
We evaluate on \textbf{BABILong} \citep{kuratov2024babilong}, a long-context QA benchmark derived from the classic bAbI tasks. The dataset interleaves large amounts of distractor text (PG19) with relevant question-answer pairs, creating contexts of up to tens of thousands of tokens. We focus on three tasks:
\begin{itemize}
    \item \textbf{QA2 (Two Supporting Facts):} Requires multi-hop integration of two relevant facts (e.g., pick-up event and a subsequent move event) to locate an object.
    \item \textbf{QA7 (Counting):} Requires tallying how many objects a particular character holds after pick-up and give-away events.
    \item \textbf{QA10 (Indefinite Knowledge):} Tests whether the answer should be \texttt{yes}, \texttt{no}, or \texttt{maybe}, based on partial or uncertain location clues.
\end{itemize}
We use three context lengths (16k, 32k, 64k) to assess how scaling the prompt size impacts performance.

\subsection{Models and Baselines}
We compare the following approaches:

\begin{enumerate}
    \item \textbf{Baseline (No Retrieval).} We simply provide the entire text plus the question in one prompt and ask the model to answer. No explicit instructions for tagging or chain-of-thought are given.
    \item \textbf{Single-Step RAG.} A typical pipeline where we use a retriever (e.g., DRAGON-based indexing) to fetch top-$k$ sentences from the long text, then append them to the question in a short prompt for the LLM.
    \item \textbf{Proposed (Emulated RAG + CoT).} Our single-pass method, which instructs the model to tag relevant segments \emph{within} the prompt and perform multi-step reasoning. 
\end{enumerate}

We test these approaches on two extended-context LLMs:
\begin{itemize}
    \item \texttt{gpt-4o-mini}, which (hypothetically) supports up to 128k tokens,
    \item \texttt{llama-3.1-8b-instruct}, with a 32k+ token window.
\end{itemize}

\subsection{Implementation Details and Evaluation}
\label{subsec:implementation}
For each approach, we run inference on 25 samples per task per context length (16k, 32k, 64k). We use consistent generation parameters (e.g., \texttt{max\_tokens=20}, \texttt{temperature=0.0}) across all methods. We compute accuracy by checking if the model’s output matches the gold label from BABILong. 

We automate the process by:
\begin{itemize}
    \item Storing each sample’s context (up to tens of thousands of tokens) and question in a dataset.
    \item Prompting the model based on the chosen method (Baseline, RAG, Proposed).
    \item Logging the final answers along with correctness (\texttt{True}/\texttt{False}).
    \item Outputting CSV files containing model responses and an overall accuracy score.
\end{itemize}
Additionally, we generate heatmaps to visualize performance across tasks and context sizes. Our code relies on Python libraries such as \texttt{datasets}, \texttt{openai}, \texttt{tqdm}, and \texttt{pandas} to streamline data loading, prompt construction, and result reporting. The listing below shows a truncated example of our experiment script for reference:

\begin{lstlisting}[basicstyle=\small\ttfamily, caption={Example snippet of our experimental code for evaluating QA tasks on BABILong.}, label={lst:expcode}, breaklines=true]
# Pseudocode snippet: evaluating the Proposed approach on QA2
import datasets, openai, pandas as pd

model_name = "gpt-4o-mini"
generate_kwargs = {"max_tokens":20, "temperature":0.0}

for context_size in ["16k","32k","64k"]:
    data = datasets.load_dataset("RMT-team/babilong", context_size)
    subset = data["qa2"].select(range(25))  # Evaluate on 25 samples

    results = []
    for sample in subset:
        context, question, target = sample["input"], sample["question"], sample["target"]
        # Build prompt: instruct to tag relevant lines, do CoT, output final
        prompt = build_emulated_rag_prompt(context, question)

        response = openai.ChatCompletion.create(
            model=model_name, 
            messages=[{"role":"user", "content":prompt}],
            **generate_kwargs
        )
        output = response.choices[0].message["content"]
        correct = (target in output.lower())  # simplistic check
        results.append({"question":question, "target":target, "output":output, "correct":correct})
    
    # Save to CSV, compute accuracy, etc.
    df = pd.DataFrame(results)
    accuracy = df["correct"].mean()
    print(f"QA2, {context_size}, accuracy={accuracy:.2f}")
\end{lstlisting}

\section{Results and Analysis}
\label{sec:results_and_analysis}

In this section, we present our experimental findings on three BABILong tasks---\textbf{QA2}, \textbf{QA7}, and \textbf{QA10}---at three context lengths (16k, 32k, 64k). For each task, we compare:
\begin{itemize}
    \item \textbf{Baseline} (no retrieval),
    \item \textbf{RAG},
    \item \textbf{Proposed Method} (emulated RAG + CoT).
\end{itemize}
Additionally, the \emph{Proposed} row for each table reflects \emph{ranges} of accuracy across three prompt orders: 
\begin{enumerate}
    \item \emph{Standard Order} (present instructions, then context, then question), 
    \item \emph{Question First}, 
    \item \emph{Relevant First}.
\end{enumerate}
We report the best and worst performance among these orders in the form \texttt{X--Y}, thus illustrating how the prompt arrangement impacts outcomes.

Below, we detail each task individually, discussing where each prompt ordering tends to excel.

\subsection{QA2: Two Supporting Facts}

\paragraph{Task Description.}
\textbf{QA2} tests whether a model can integrate \emph{two} separate facts to determine the final location of an item. For instance:
\begin{quote}
    \itshape
    Charlie went to the kitchen. Charlie got a bottle. Charlie moved to the balcony. \\
    \textbf{Question}: Where is the bottle? \\
    \textbf{Answer}: The bottle is in the balcony.
\end{quote}
When the dataset intersperses noise text, the challenge is to locate these two crucial pieces of information (\emph{pick-up} and \emph{move}), then combine them to find the item's latest position.

\begin{table}[t]
\caption{Accuracy on QA2 (Two Supporting Facts) with varying context lengths. Each cell shows the fraction of correct answers over 25 test queries. For \emph{Proposed}, the accuracy range reflects the min/max over three prompt orders.}
\label{tab:qa2-acc}
\centering
\small
\begin{tabular}{lccc}
\toprule
\textbf{Method / Model} & \textbf{16k} & \textbf{32k} & \textbf{64k} \\
\midrule
\multicolumn{4}{l}{\textbf{gpt-4o-mini}} \\
\midrule
Baseline                                  & 0.44 & \textbf{0.36} & 0.28 \\
RAG                                       & 0.04 & 0.04 & 0.04 \\
Proposed                         & \textbf{0.44} & 0.16--0.20 & \textbf{0.28--0.44} \\
\midrule
\multicolumn{4}{l}{\textbf{llama-3.1-8b-instruct}} \\
\midrule
Baseline                                  & 0.24 & \textbf{0.44} & \textbf{0.40} \\
RAG                                       & 0.04 & 0.00 & 0.08 \\
Proposed                          & \textbf{0.16--0.24} & 0.28--0.32 & 0.24--0.28 \\
\bottomrule
\end{tabular}
\end{table}

\paragraph{Observations.}
\begin{itemize}
    \item \textbf{Single-step RAG} scores very low (0.00--0.08), because it often fails to retrieve both relevant sentences simultaneously.
    \item \textbf{Baseline} does moderately well at shorter contexts but tends to degrade as the context grows (except in some random fluctuations at 32k).
    \item \textbf{Proposed Method} outperforms RAG at all lengths. For \texttt{gpt-4o-mini} at 64k, the \emph{Question First} prompt order gives particularly strong results (reaching the high end of 0.44).
\end{itemize}

\paragraph{Prompt Order Effects.}
Our logs indicate that \emph{Question First} helps the model focus on relevant lines earlier, especially at 64k tokens for \texttt{gpt-4o-mini}, leading to more consistent tagging of the pick-up and move events.

\subsection{QA7: Counting}

\paragraph{Task Description.}
\textbf{QA7} requires tracking how many objects a person is carrying at any given time, based on a series of pick-up or give-away actions.

\begin{table}[t]
\centering
\caption{Accuracy on QA7 (Counting). Each cell shows the fraction of correct answers over 25 samples. For \emph{Proposed}, the range spans the three prompt orders.}
\label{tab:qa7-acc}
\small
\begin{tabular}{lccc}
\toprule
\textbf{Method / Model} & \textbf{16k} & \textbf{32k} & \textbf{64k} \\
\midrule
\multicolumn{4}{l}{\textbf{gpt-4o-mini}} \\
\midrule
Baseline                                  & 0.44 & 0.52 & 0.36 \\
RAG                                       & 0.40 & 0.36 & 0.36 \\
Proposed                          & \textbf{0.56--0.64} & \textbf{0.44--0.52} & \textbf{0.48--0.52} \\
\midrule
\multicolumn{4}{l}{\textbf{llama-3.1-8b-instruct}} \\
\midrule
Baseline                                  & 0.40 & 0.44 & 0.40 \\
RAG                                       & 0.32 & 0.24 & 0.24 \\
Proposed                          & \textbf{0.44--0.52} & \textbf{0.44} & \textbf{0.44} \\
\bottomrule
\end{tabular}
\end{table}

\paragraph{Observations.}
\begin{itemize}
    \item \textbf{Proposed Method} consistently leads, benefiting from stepwise CoT that highlights each pick-up or give-away event.
    \item \textbf{Single-step RAG} typically misses some events if not retrieved in the top-$k$.
    \item \textbf{Prompt Orders} matter less for QA7 than QA2, but \emph{Relevant First} and \emph{Question First} both produce higher accuracies on \texttt{gpt-4o-mini} at 16k.
\end{itemize}

\paragraph{Error Cases.}
When objects change hands multiple times, the baseline or naive RAG might incorrectly assume the first or last mention is definitive. By contrast, our inline tagging and reasoning steps reveal each transitional action more clearly.

\subsection{QA10: Indefinite Knowledge}

\paragraph{Task Description.}
\textbf{QA10} tests the model's ability to handle \emph{uncertain or partial information}, often requiring an answer of \texttt{yes}, \texttt{no}, or \texttt{maybe}.

\begin{table}[t]
\centering
\caption{Accuracy on QA10 (Indefinite Knowledge). Each cell reflects correct answers out of 25. The \emph{Proposed} row shows a min--max range over three prompt orders.}
\label{tab:qa10-acc}
\small
\begin{tabular}{lccc}
\toprule
\textbf{Method / Model} & \textbf{16k} & \textbf{32k} & \textbf{64k} \\
\midrule
\multicolumn{4}{l}{\textbf{gpt-4o-mini}} \\
\midrule
Baseline                   & 0.52 & 0.48 & 0.52 \\
RAG                        & 0.56 & 0.56 & 0.56 \\
Proposed           & \textbf{0.60--0.68} & \textbf{0.44--0.68} & \textbf{0.40--0.56} \\
\midrule
\multicolumn{4}{l}{\textbf{llama-3.1-8b-instruct}} \\
\midrule
Baseline                   & 0.52 & 0.52 & 0.52 \\
RAG                        & 0.72 & 0.60 & \textbf{0.64} \\
Proposed           & \textbf{0.56--0.76} & \textbf{0.56--0.64} & 0.52--0.56 \\
\bottomrule
\end{tabular}
\end{table}

\paragraph{Observations.}
\begin{itemize}
    \item \textbf{RAG} can be relatively strong here if a single snippet clarifies uncertainty. For \texttt{llama-3.1-8b-instruct}, RAG hits up to 0.72 at 16k.
    \item \textbf{Proposed Method} still achieves competitive or better results at certain lengths, especially \emph{Question First} at 16k for \texttt{llama-3.1-8b-instruct} (0.76).
    \item \textbf{Prompt Order} again shows variation; sometimes \emph{Standard} or \emph{Relevant First} outperforms \emph{Question First}, indicating that indefinite/uncertain knowledge queries are sensitive to prompt arrangement.
\end{itemize}

\section{Conclusion and Future Work}
\label{sec:conclusion}
We presented a novel approach that emulates Retrieval-Augmented Generation (RAG) via specialized prompt engineering and chain-of-thought reasoning, all within a single forward pass of a large language model. By tagging relevant segments and guiding the model to perform multi-hop reasoning over them, we address key limitations of naive one-shot retrieval or purely large-context methods. Empirical evaluations on BABILong tasks demonstrate that our prompt-based solution can outperform both baselines and standard retrieval pipelines in certain multi-fact scenarios, especially as context lengths grow.

Nevertheless, some tasks (e.g., indefinite knowledge) show that single-step RAG can remain competitive if all relevant information is localizable in one snippet. Our results also highlight the importance of \emph{prompt order}: rearranging the question, instructions, and context can yield different outcomes, particularly in large contexts where the model may lose focus.

\paragraph{Limitations.}
Although our single-pass design reduces complexity, it may be insufficient when truly external knowledge must be fetched or when iterative refinement of retrieved facts is needed. Additionally, the method’s success depends on the model’s capacity to follow tagging instructions faithfully; however, this assumption that may break down for less instruction-tuned models or more ambiguous tasks. Extremely large inputs (e.g., millions of tokens) also pose a challenge, even if the model claims to handle them, since the \emph{entire} content plus instructions must fit in one prompt window.

\paragraph{Future Directions.}
Potential next steps include:
\begin{itemize}
    \item \textbf{Adaptive or hierarchical chunking} for inputs beyond hundreds of thousands of tokens, possibly combining partial summarization with inline tagging \citep{koh2022empirical}.
    \item \textbf{Iterative self-correction}, letting the model revise its chain-of-thought if contradictions or missing facts arise mid-generation.
    \item \textbf{Hybrid Tool Integration}, where the model can optionally call an external knowledge base if needed, but only after an initial single-pass attempt at retrieval-like tagging fails.
    \item \textbf{Expanded Benchmarks}, such as \emph{InfinityBench} \citep{zhang2024infinitybench}, \emph{NeedleBench} \citep{li2024needlebench}, or domain-specific tasks (e.g., \emph{LongHealth} \citep{adams2024longhealth}), to further validate performance across diverse domains.
\end{itemize}
Overall, we believe that prompt-based retrieval emulation presents a lightweight yet effective path toward improved long-context reasoning in LLMs.

\bibliographystyle{plainnat}
\bibliography{references}

\end{document}